\documentclass[10pt,conference,a4paper]{IEEEtran}
\usepackage{cite}

\ifCLASSINFOpdf
  \usepackage[pdftex]{graphicx}
  \DeclareGraphicsExtensions{.pdf,.jpeg,.png}
\else
\fi
\begin{document}
%
\title{Do We Really Need Scene-specific Pose Encoders?}

\author{\IEEEauthorblockN{Yoli Shavit}
\IEEEauthorblockA{Toga Networks a Huawei Company\\
Email: yoli.shavit@huawei.com}
\and
\IEEEauthorblockN{Ron Ferens}
\IEEEauthorblockA{Toga Networks a Huawei Company\\
Email: ron.ferens@huawei.com}}


%


\maketitle

\let\thefootnote\relax\footnote{© 2020 IEEE. Personal use of this material is permitted. Permission from IEEE must be obtained for all other uses, in any current or future media, including reprinting/republishing this material for advertising or promotional purposes, creating new collective works, for resale or redistribution to servers or lists, or reuse of any copyrighted component of this work in other works.}

\begin{abstract}
Visual pose regression models estimate the camera pose from a query image with a single forward pass. Current models learn pose encoding from an image using deep convolutional networks which are trained per scene. The resulting encoding is typically passed to a multi-layer perceptron in order to regress the pose. In this work, we propose that scene-specific pose encoders are not required for pose regression and that encodings trained for visual similarity can be used instead. In order to test our hypothesis, we take a shallow architecture of several fully connected layers and train it with pre-computed encodings from a generic image retrieval model. We find that these encodings are not only sufficient to regress the camera pose, but that, when provided to a branching fully connected architecture, a trained model can achieve competitive results and even surpass current \textit{state-of-the-art} pose regressors in some cases. Moreover, we show that for outdoor localization, the proposed architecture is the only pose regressor, to date, consistently localizing in under 2 meters and 5 degrees.
\end{abstract}


%
\IEEEpeerreviewmaketitle

\section{Introduction}
Solving the pose estimation problem amounts to accurately predicting the position and orientation of an agent within a predefined coordinate system. The agent can be a robot, a handheld device or a vehicle. In visual pose estimation, we aim to solve this challenge solely using visual inputs generated by RGB and/or depth sensors mounted on the agent. That is, we assume that given enough visual clues, one can deduce the pose within the reference scene. 
State-of-the-art methods for visual pose estimation such as \cite{sattler2016efficient,taira2018inloc,sarlin2019coarse}, rely on the availability of a structural model that can map 2D points in an image to 3D points in the reference scene (structure-based methods). Given 2D-2D matches between the query image and one or more train images, 2D-3D matches are obtained using the structural model, and passed to a Perspective-n-Point (PnP) algorithm in conjunction with RANSAC \cite{fischler1981random}. The resulting estimated pose/s can be further evaluated and re-ranked with a pose verification procedure \cite{taira2018inloc,taira2019right}. In the past few years, deep learning methods for regressing the camera pose from a single image have emerged \cite{kendall2015posenet,kendall2016modelling,kendall2017geometric,melekhov2017image,naseer2017deep,walch2017image,wu2017delving,Cai2018AHP}.  While these absolute pose regressors (APRs) are typically an order of magnitude  less accurate than their structure-based counterpart methods and require scene specific training, they  offer a significant reduction in runtime (milliseconds versus seconds) and replace the heavy structure-based pipeline with a single forward pass.  

\begin{figure}[!t]
	\centering
	\includegraphics[width=\linewidth]{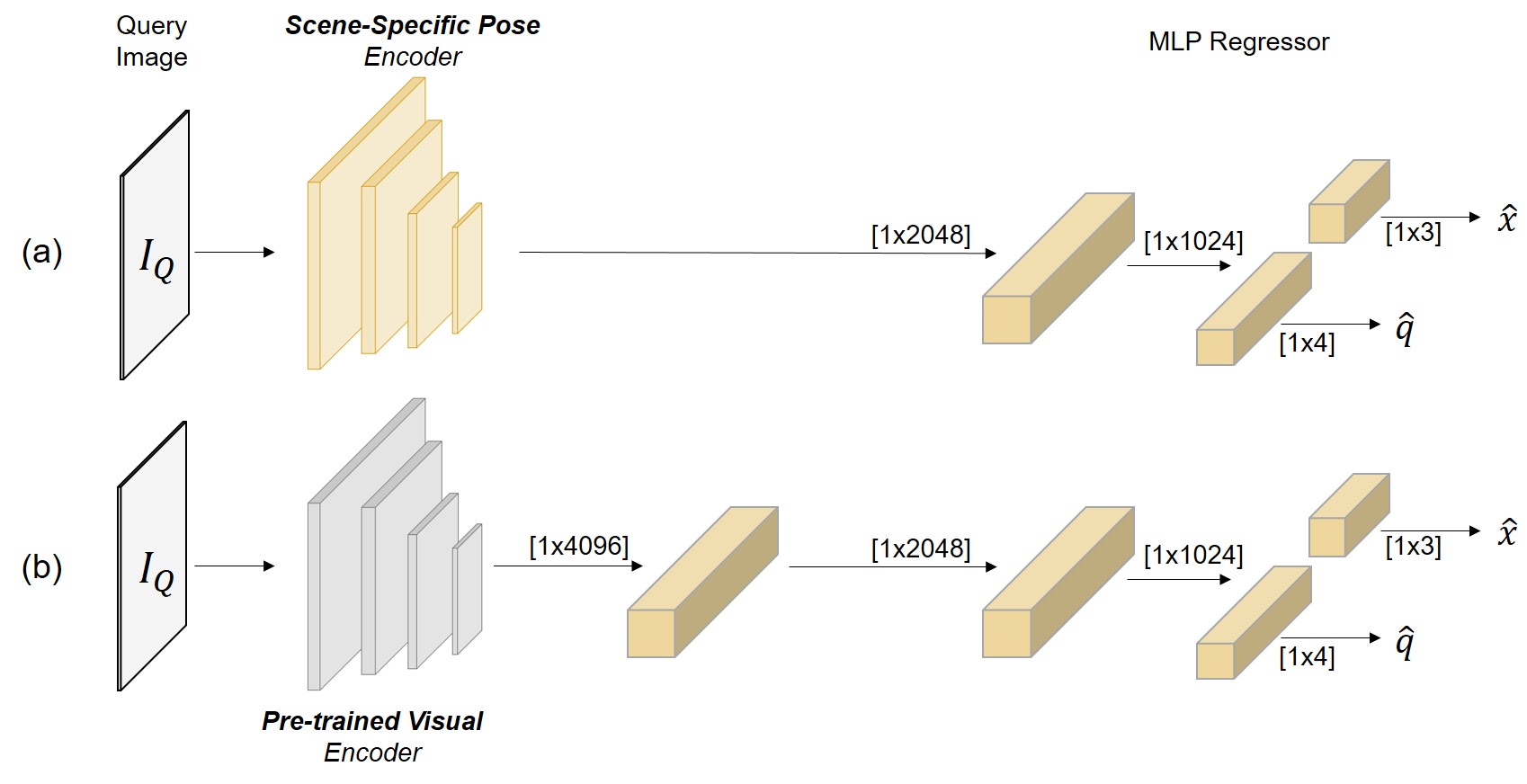}
	\caption{End-to-end pose regression architectures. (a) A representative architecture of a pose repressor, using a scene-specific pose encoder and an MLP regressor. (b) An MLP architecture using  generic scene-independent encodings, learnt for place recognition. Yellow coloring indicates layers that are updated during training. }
	\label{fig1_aprs}
\end{figure}

A representative architecture for APRs consists of a convolutional pose encoder, based on a common backbone architectures such as GoogLeNet  \cite{szegedy2015going} or ResNet \cite{he2016deep}, followed by a multilayer perceptron (MLP) for separately regressing the position and orientation vectors (Fig. \ref{fig1_aprs}a). The pose encoder is commonly initialized using weights trained for object classification or place recognition and then fine tuned as part of the training process for encoding the training set of the scene of interest.

Recently, a comprehensive study of APRs \cite{sattler2019understanding} framed camera pose regression methods as an approximation of image retrieval (IR), where the network learns some interpolation of memorized images and their poses from the training set. The authors further showed that APRs cannot consistently outperform an IR based method, which estimates the pose through interpolation of nearest neighbors' poses.  Following this work, we ask whether scene- (or even dataset-) specific pose encoders are actually necessary for performing pose regression? To answer this question, we take an MLP architecture typically used by APRs (Fig. \ref{fig1_aprs}a). We train it with encodings generated using an IR model which was pre-trained to capture visual similarity for place recognition. As oppose to other APRs, here the encoder is not involved in the training process and the images' encodings, rather than the images themselves, are used as input for learning the camera pose regression task (Fig. \ref{fig1_aprs}b). We find that the resulting model achieves similar performance to a baseline pose regressor \cite{kendall2015posenet} (see Table \ref{tb:baseline_cambridge_res}). 

\begin{figure*}[t]
	\centering
	\includegraphics[scale=0.5]{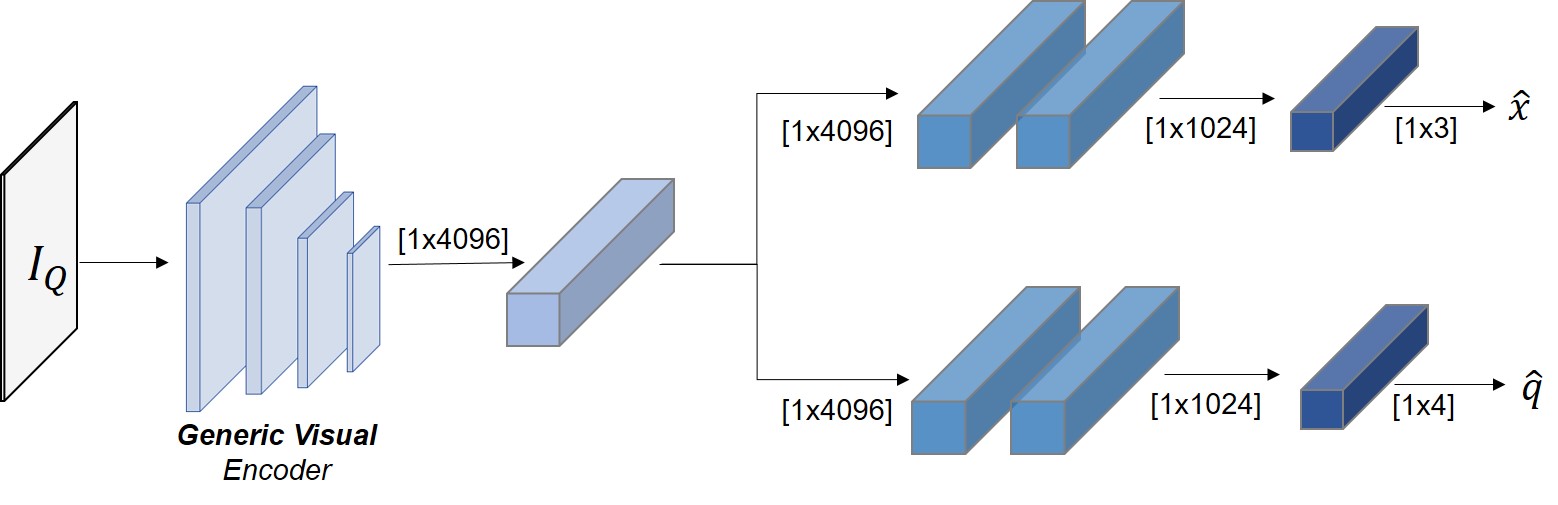}
	\caption{IRPNet architecture. An MLP with two branches for separately regressing the position and orientation vectors }
	\label{fig2_irpnet}
\end{figure*}

Secondly, we extend the above MLP with a branching architecture (Fig. \ref{fig2_irpnet}), similarly to \cite{naseer2017deep}. We name this architecture \textit{IRPNet}, as it uses IR encodings for pose regression. In order to teach our model to regress the camera pose, the loss function needs to be optimized for both position and orientation objectives, which differ in scale. This multi-objective tradeoff is challenging as the network needs to optimize for related, yet different tasks in nature. A common solution is to introduce a balancing factor, either manually \cite{kendall2015posenet} or through learning \cite{kendall2017geometric}. However, they still require dataset-specific fine-tuning \cite{kendall2015posenet} or initialization \cite{valada2018deep}. Instead, we train separately for orientation and position.

We evaluate IRPNet on outdoor and indoor datasets (Tables \ref{tb:cambridge_res} and \ref{tb:7scenes_res}, respectively), commonly used to benchmark APRs. We find that IRPNet consistently localizes in under 2$m$ and 5$^\circ$ on an outdoor benchmark. This lower bound cannot be guaranteed by any other APR, to date, or by an IR based interpolation method proposed by \cite{sattler2019understanding}. In addition, together with another \textit{state-of-the-art} APR, IRPNet is the only model able to rank among the top-three APRs on both outdoor and indoor dataets (Tables \ref{tb:cambridge_rank} and \ref{tb:7scenes_rank}, respectively). IRPNet also offers several other advantages compared to APRs with pose encoders: it is faster to train (10x) and does not require scaling or cropping of the query image. It also facilitates lighter storage of the train dataset as image encodings can be stored instead of the images themselves. Compared to scene-specific encodings, an independent model allows extending the train set without retraining the encoder and elevates the need to pre-compute encodings more than once.

To summarize, our contributions are as follows:
\begin{itemize}
	\item We propose a new paradigm for camera pose regression, where scene-specific encoders are not required and scene-agnostic IR encoders can be used instead.
	\item We test our hypothesis and show that encodings trained for visual similarity are sufficient for learning to regress the camera pose.
	\item We introduce IRPNet, a branching MLP architecture, which consistently rank among the top-three APRs on both outdoor and indoor datasets and is the only model to provide a consistent lower bound on orientation and position errors on an outdoor benchmark. In addition, the proposed method enables lighter database storage, is faster to train and does not require image pre-processing such as scaling or cropping.  
\end{itemize}

\section{Related Work}
\subsection{Absolute Pose Regression}
Visual pose regression was first suggested by \cite{kendall2015posenet}. The proposed architecture, named PoseNet, was based on a GoogLeNet backbone where classifiers were replaced with  fully-connected (FC) layers to regress the absolute pose of the input image. The model was trained to minimize a Pose Loss function for an image $I$:
\begin{equation}\label{eq1_pose_loss}
L(I) = s_{x} L_{x}(I) + s_{q} L_{q}(I) 
\end{equation}
$L_{x}(I)$, the Position Loss, is the Euclidean distance between the ground truth and estimated position vectors: $L_{x}(I) = ||\hat{x}-x||_{2}$. $L_{q}(I)$, the Orientation Loss, is the Euclidean distance between the ground truth and estimated unit vector quaternions:   $L_{q}(I) = ||\hat{q}-\frac{q}{||q||}||_{2}$. The hyperparameters $ s_{x}$ and $s_{q}$ control the tradeoff between the two losses. In the original work, $ s_{x}$ was  set to $1$ and $s_{q}$ (referred to as $\beta$) was fine-tuned for each scene and differed in scale between outdoor and indoor scenes. 

The proposed layout of a deep convolutional encoder followed by an MLP regressor, was soon adopted as a baseline for further improvements. Different variations to the encoder and MLP architectures were proposed in order to reduce overfitting \cite{kendall2016modelling,kendall2017geometric,melekhov2017image,walch2017image,wu2017delving}. For example, replacing the GoogLeNet encoder with a ResNet encoder \cite{melekhov2017image} or duplicating the MLP components to form a branching architecture \cite{wu2017delving,naseer2017deep}. 
In addition, extensions to the loss function were proposed in order to better learn the balance between the orientation and position objectives \cite{kendall2017geometric} or to incorporate other modalities \cite{brahmbhatt2018geometry}. For a detailed review of architecture families and losses, used for camera pose regression, see \cite{shavit2019introduction}. 

In terms of accuracy, APRs are currently inferior to structure-based methods \cite{sattler2019understanding}. However, they only require an RGB input, have a small memory signature and an order of magnitude shorter inference time \cite{kendall2015posenet}. More recently, researchers have proposed to jointly learn absolute pose regression with other tasks such as relative pose regression (RPR, finding the relative motion between two images) and semantic segmentation \cite{valada2018deep,radwan2018vlocnet++}. Learning these additional auxiliary tasks led to a significant improvement in performance over other pose regressors, matching classical \textit{state-of-the-art} pipelines in some cases. However, the learning was designed for a sequential workflow (relying on the previous pose to predict the current pose), making the proposed approach less suitable for re-localization scenarios. 

In order to avoid the need to train a model per scene, some methods focused on improving the generalization of pose regressors \cite{balntas2018relocnet,laskar2017camera,ding2019camnet}. These models were first trained offline for RPR on a reference dataset. The trained encoder was then used to fetch the nearest neighbor/s in order to estimate the relative motion to the query. Using the known pose of the neighbor/s and the estimated relative motion, the pose of the query was finally estimated. While RPR-based methods improved generalization of absolute pose regression, their pose encoders were still specific to a given dataset and the generated encodings could not be transfered between different datasets. In addition, generalization either came on the expense of accuracy \cite{balntas2018relocnet,laskar2017camera}, or when high accuracy was achieved, it was only evaluated on a small scale indoor dataset \cite{ding2019camnet}.


\subsection{Image Retrieval}
Image retrieval methods rely on global image representations to efficiently index and fetch images. These global descriptors can be computed by aggregating local features using methods such as VLAD \cite{jegou2010aggregating} and Fisher Vector \cite{jegou2011aggregating}. More recently, deep convolutional neural networks (CNNs) were proposed as a surrogate for aggregation methods \cite{arandjelovic2016netvlad,noh2017large,dusmanu2019d2}. For example, NetVLAD \cite{arandjelovic2016netvlad}, fine-tunes a pretrained CNN encoder, followed by a learnt VLAD component (convolution and soft assignment), PCA and whitening. Learned global descriptors were shown to be useful for place recognition, achieving \textit{state-of-the-art} localization \cite{arandjelovic2016netvlad,noh2017large,dusmanu2019d2}. 

In the context of visual pose estimation, image retrieval methods can provide a rough estimate by using the known poses of the query’s  nearest neighbors. For example, by taking the pose of the closest neighbor. As part of studying the properties of APRs, a recent work  \cite{sattler2019understanding} proposed an IR based baseline, where the pose of a query image is estimated with a weighted sum of its nearest neighbors's poses. The weighting is found at inference time, by minimizing the distance between the visual encodings of the neighbors' and the query image. The performance of this baseline was not consistently surpassed on indoor and outdoor benchmarks by any APR, to date.

Visual pose estimation pipelines \cite{taira2018inloc,sarlin2019coarse}, achieving \textit{state-of-the-art} accuracy on large-scale datasets with challenging conditions, also leverage on IR encodings. Pretrained models for place recognition are used to fetch a relatively small set of nearest neighbors, providing a reduced search space for subsequent feature matching. In addition, latent feature maps, generated by IR models, were shown to effectively replace classical features in the matching stage \cite{taira2018inloc,noh2017large,dusmanu2019d2}.

\section{Regressing camera pose without a pose encoder}

\subsection{Image retrieval encoder for camera pose estimation}\label{IR_encode_for_pose}
\begin{figure*}[h!]
	\centering
	\includegraphics[scale=0.5]{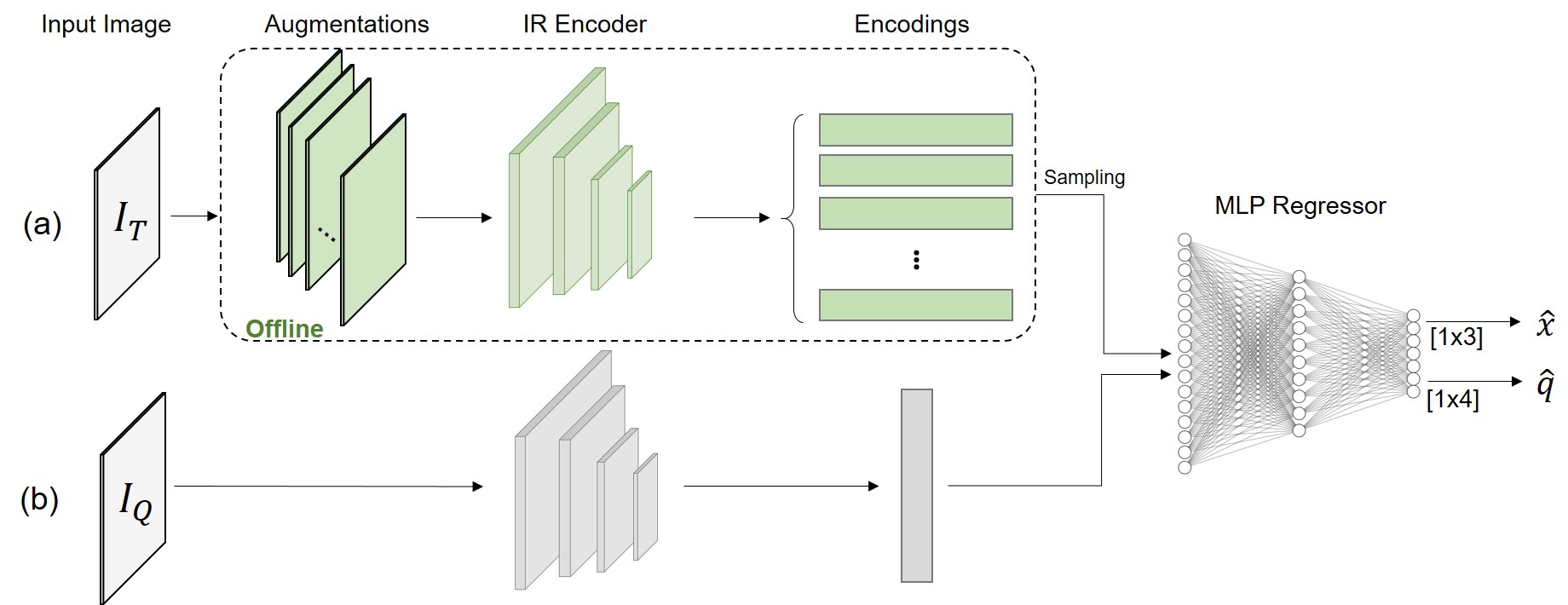}
	\caption{Our proposed training (a) and testing (b) procedure for camera pose regression using IR encoding. (a) At train time we sample a single entry from pre-computed IR encodings of the input image ($I_T$) augmentations and feed it to an MLP regressor. (b) At test time, given a query image ($I_Q$), we generate its IR encoding and infer the camera pose by passing it to through an MLP regressor.}
	\label{fig3_train_aug}
\end{figure*}

Although IR visual encodings have proven to be useful for place recognition and pose estimation pipelines, they have not been exploited by pose regression methods. Here, we propose to remove the pose encoder and use instead global descriptors learned for place recognition. 
In order to test this paradigm we take an MLP architecture, commonly used by APRs (Fig. \ref{fig1_aprs}a), and add one FC layer to match the input dimension of the MLP. The resulting architecture (Fig. \ref{fig1_aprs}b), consists  of a sequence of 4096-, 2048- and 1024-dimensional FC layers, followed by a 3-dimensional FC layer for regressing the position and a 4-dimensional FC layer for regressing the orientation. In order to generate the input for our model, we apply a NetVLAD model with a VGG16 \cite{simonyan2014very} backbone, pretrained on the Pittsburgh 250K dataset \cite{torii2013visual}. We generate $m$ augmented encodings for each image by randomly jittering its color, hue and saturation $m$ times and then encoding the resulting augmentations. At train time, we randomly select an augmentation and feed it to the network (Fig. \ref{fig3_train_aug}a). At test time, the query image is loaded and its encoding is computed on-the-fly (Fig. \ref{fig3_train_aug}b). We normalize images using the mean and standard deviation used in \cite{arandjelovic2016netvlad}, but no further range scaling, resizing or cropping is applied at train or at test time.
We train our model with a commonly used pose loss \cite{kendall2017geometric} where the scaling between the position and orientation objectives is learned as part of the training process.

Table \ref{tb:baseline_cambridge_res} shows the results when training on an outdoor dataset (Cambridge Landmarks \cite{kendall2015posenet}). Further details about the training procedure and the datasets used for evaluation are provided in Section \ref{experiments}. Although we do not perform any fine-tuning and use a pre-trained encoder, the shallow MLP model is able to outperform the baseline APR in $62.5\%$ of the error measurements and consistently achieves a significantly smaller orientation error (1.8x on average).

\begin{table}[h!]
	\centering
	\caption{Median position error (in meters) and orientation error (in degrees) of a baseline APR \cite{kendall2015posenet} and a representative MLP architecture trained with visual encodings.}
	\label{tb:baseline_cambridge_res}
	\begin{tabular}{|c||c|c|c|c|c|}
		\hline
		Method & K. College & Old Hospital & Shop Facade & St. Mary \\	
		\hline
		PoseNet & \textbf{1.92}$/5.4$ & \textbf{2.31}$/5.38$ & $1.46/8.08$ & \textbf{2.65}$/8.48$ \\
		\hline
		MLP & $1.96/$\textbf{2.53} & $2.62/$\textbf{4.00} & \textbf{1.35}/\textbf{3.70} & $2.84/$\textbf{5.79} \\
		\hline
	\end{tabular}
\end{table}

\subsection{Network Architecture}
Encouraged by the initial proof of concept demonstrated in section  \ref{IR_encode_for_pose}, we propose IRPNet, an MLP architecture for regressing camera pose from IR encodings. We follow previous works \cite{naseer2017deep,wu2017delving} and split  the MLP into two branches (one per objective) and add one 4096-dimensional shared layer. Each branch consists of a sequence of a 4096-, 2048-, 1024- and $p$-dimensional FC layers. One branch regresses the position vector ($p=3$) and one branch regresses the orientation vector ($p=4$). The architecture of IRPNet is shown is Fig \ref{fig2_irpnet}.

\begin{table*}[ht]
	\centering
	\caption{Median position error (in meters) and orientation error (in degrees) of APRs, when tested on the Cambridge dataset.}
	\label{tb:cambridge_res}
	\begin{tabular}{|c||c|c|c|c|c|}
		\hline
		Method & K. College & Old Hospital & Shop Facade & St. Mary & Average	\\	
		\hline
		PoseNet (Learnable)\cite{kendall2017geometric} & $0.99/1.06$ & $2.17/2.94$ & $1.05/3.97$ & $1.49/3.43$ & $1.43/2.85$ \\
		\hline
		GeoPoseNet  \cite{kendall2017geometric} & $0.88/1.04$ & $3.20/3.29$ & $0.88/3.78$ & $1.57/3.32$ & $1.63/2.86$ \\
		\hline
		\textbf{IPRNet (Ours)} & $1.18/2.19$ & $1.87/3.38$ & $0.72/3.47$ & $1.87/4.94$ & $1.42/3.45$ \\
		\hline
		MapNet \cite{brahmbhatt2018geometry} & $1.07/1.89$ & $1.94/3.91$ & $1.49/4.22$ & $2.0/4.53$ & $1.63/3.64$ \\
		\hline
		LSTM-PN \cite{walch2017image} & $0.99/3.65$ & $1.51/4.29$ & $1.18/7.44$ & $1.52/6.68$ & $1.30/5.52$ \\
		\hline
		GPoseNet & $1.61/2.29$ & $2.62/3.89$ & $1.14/5.73$ & $2.93/6.46$ & $2.08/4.59$ \\
		\hline
		BayesianPN \cite{kendall2016modelling} & $1.74/4.06$ & $2.57/5.14$ & $1.25/7.54$ & $2.11/8.38$ & $1.92/6.28$ \\
		\hline
		PoseNet \cite{kendall2015posenet} & $1.92/5.4$ & $2.31/5.38$ & $1.46/8.08$ & $2.65/8.48$ & $2.09/6.84$ \\
		\hline
		SVS-Pose \cite{naseer2017deep} & $1.06/2.81$ & $1.50/4.03$ & $0.63/5.73$ & $2.11/8.11$ & $1.33/5.17$ \\
		\hline
		IR Baseline \cite{sattler2019understanding} & $1.48/4.45$ & $2.68/4.63$ & $0.9/4.32$ & $1.62/6.06$ & $1.67/4.87$ \\
		\hline
	\end{tabular}
\end{table*}

\begin{table*}[ht]
	\centering
	\caption{Position, orientation rankings of APRs, when tested on the Cambridge dataset. The final ranking is taken as the average between the average position rank and average orientation rank.}
	\label{tb:cambridge_rank}
	\begin{tabular}{|c||c|c|c|c|c|c|}
		\hline
		Method & K. College & Old Hospital & Shop Facade & St. Mary & Average	& Final Rank \\	
		\hline
		PoseNet (Learnable) & $2, 2$  & $5, 1$ & $4, 3$ & $1, 2$ & $3, 2$ & $1$ \\
		\hline
		GeoPoseNet & $1, 1$ & $9, 2$ & $3, 2$ & $3, 1$ & $4, 2$ & $2$ \\
		\hline
		\textbf{IPRNet (Ours)} & $6, 4$  & $3, 3$ & $2$, \textbf{1} & $4, 4$ & $4, 3$ & \textbf{3} \\
		\hline
		MapNet & $5, 3$ & $4, 5$ & $9, 4$ & $5, 3$ & $6, 4$ & $6$ \\
		\hline
		LSTM-PN & $2, 7$ & $2, 7$ & $6, 7$ & $2, 6$ & $1, 7$ & $5$ \\
		\hline
		GPoseNet & $7, 5$ & $8, 4$ & $5, 5$ & $9, 5$ & $7, 5$ & $7$ \\
		\hline
		BayesianPN & $8, 8$ & $7, 8$ & $7, 8$ & $6, 8$ & $7, 8$ & $8$ \\
		\hline
		PoseNet & $9, 9$ & $6, 9$ & $8, 9$ & $8, 9$ & $8, 9$ & $9$ \\
		\hline
		SVS-Pose & $4, 6$ & $1, 6$ & $1, 5$ & $6, 7$ & $3, 6$ & $3$ \\
		\hline
	\end{tabular}
\end{table*}

\begin{table*}[ht]
	\centering
	\caption{Median position error (in meters) and orientation error (in degrees) of APRs, when tested on the 7 Scenes dataset.}
	\label{tb:7scenes_res}
	\begin{tabular}{|c||c|c|c|c|c|c|c|c|}
		\hline
		Method & Chess & Fire & Heads & Office & Pumpkin & Kitchen & Stairs & Average \\	
		\hline
		PoseNet (Learnable) & $0.14/4.50$ & $0.27/11.8$ & $0.18/12.1$ & $0.2/5.77$ & $0.25/4.82$ & $0.24/5.52$ & $0.37/10.6$ & $0.24/7.87$ \\
		\hline
		GeoPoseNet & $0.13/4.48$ & $0.27/11.3$ & $0.17/13.0$ & $0.19/5.55$ & $0.26/4.75$ & $0.23/5.35$ & $0.35/12.4$ & $0.23/8.12$ \\
		\hline
		\textbf{IPRNet (Ours)} & $0.13/5.64$ & $0.25/9.67$ & $0.15/13.10$ & $0.24/6.33$ & $0.22/5.78$ & $0.30/7.29$ & $0.34/11.6$ & $0.23/8.49$ \\
		\hline
		MapNet & $0.08/3.25$ & $0.27/11.7$ & $0.18/13.3$ & $0.17/5.15$ & $0.22/4.02$ & $0.23/4.93$ & $0.3/12.1$ & $0.21/7.79$ \\
		\hline
		LSTM-PN & $0.24/5.77$ & $0.34/11.9$ & $0.21/13.7$ & $0.30/8.08$ & $0.33/7.0$ & $0.37/8.83$ & $0.40/13.7$ & $0.31/9.86$ \\
		\hline
		GPoseNet & $0.20/7.11$ & $0.38/12.3$ & $0.21/13.8$ & $0.28/8.83$ & $0.37/6.94$ & $0.35/8.15$ & $0.37/12.5$ & $0.31/9.95$ \\
		\hline
		BayesianPN & $0.37/7.24$ & $0.43/13.7$ & $0.31/12.0$ & $0.48/8.04$ & $0.61/7.08$ & $0.58/7.54$ & $0.48/13.1$ & $0.47/9.81$ \\
		\hline
		PoseNet & $0.32/8.12$ & $0.47/14.4$ & $0.29/12.0$ & $0.48/7.68$ & $0.47/8.42$ & $0.59/8.64$ & $0.47/13.8$ & $0.44/10.43$ \\
		\hline
		IR Baseline & $0.18/10.0$ & $0.33/12.4$ & $0.15/14.3$ & $0.25/10.1$ & $0.26/9.42$ & $0.27/11.1$ & $0.24/14.7$ & $0.24, 11.72$ \\
		\hline
	\end{tabular}
\end{table*}

\begin{table*}[ht]
	\centering
	\caption{Position, orientation rankings of APRs, when tested on the 7 Scenes dataset. The final ranking is taken as the average between the average position rank and average orientation rank.}
	\label{tb:7scenes_rank}
	\begin{tabular}{|c||c|c|c|c|c|c|c|c|c|}
		\hline
		Method & Chess & Fire & Heads & Office & Pumpkin & Kitchen & Stairs & Average & Final Rank \\	
		\hline
		PoseNet (Learnable) & $4, 3$  & $2, 4$ & $3, 3$ & $3, 3$ & $3, 3$ & $3, 3$ & $3, 4$ & $4, 1$ & $4$\\
		\hline
		GeoPoseNet & $2, 2$  & $2, 2$ & $2, 4$ & $2, 2$ & $4, 2$ & $1, 2$ & $3, 4$ & $3, 2$ & $2$\\
		\hline
		\textbf{IPRNet (Ours)} & $2, 4$  & \textbf{1, 1} & \textbf{1}, $5$ & $4, 4$ & \textbf{1}, $4$ & $4, 4$ & $2, 2$ & $2, 4$ & \textbf{3}\\
		\hline
		MapNet & $1, 1$  & $2, 3$ & $3, 6$ & $1, 1$ & $1, 1$ & $1, 1$ & $1, 3$ & $1, 1$ & $1$\\
		\hline
		LSTM-PN & $6, 5$  & $5, 5$ & $5, 7$ & $6, 7$ & $5, 6$ & $6, 6$ & $8, 6$ & $6, 7$ & $6$\\
		\hline
		GPoseNet & $5, 6$  & $6, 6$ & $5, 8$ & $5, 8$ & $6, 5$ & $5, 6$ & $4, 5$ & $5, 6$ & $5$\\
		\hline
		BayesianPN & $8, 7$  & $7, 7$ & $8, 1$ & $7, 6$ & $8, 7$ & $7, 5$ & $8, 6$ & $8, 5$ & $6$\\
		\hline
		PoseNet & $7, 8$  & $8, 8$ & $7, 1$ & $7, 5$ & $7, 8$ & $8, 7$ & $7, 8$ & $7, 7$ & $8$\\
		\hline
	\end{tabular}
\end{table*}

\section{Experiments}\label{experiments}

\subsection{Datasets}
For evaluation purposes we chose an indoor and an outdoor dataset, commonly used for benchmarking APRs: 7 Scenes \cite{glocker2013real} and Cambridge Landmarks \cite{kendall2015posenet}. These two datasets encompass common challenges and scenarios typical of visual localization applications: large, medium and small scales, indoor/outdoor, repeating elements, textureless features,  significant viewpoint changes and trajectory variations between train and test sets. The 7 Scenes dataset includes seven small-scale indoor scenes, with a spatial extent of a few squared meters. The Cambridge Landmarks dataset is a mid-scale urban outdoor dataset. We report results on four of its six scenes as the other two scenes were either not consistently covered in the literature or reported as possibly faulty \cite{naseer2017deep,walch2017image}.

\subsection{Training procedure}
The input to our network are encodings of augmented train images, generated as described in section \ref{IR_encode_for_pose}, with ($m=$) 10 augmentations per image. We train our network  with a mini-batch of size 8 and optimze using Adam, with $\beta_{1}=0.9$, $\beta_{2}=0.999$ and $\epsilon=10^{-10}$. 
We use an initial learning rate of $\lambda=10^{-2}$ ($\lambda=10^{-3}$) and a  weight decay of $\frac{\lambda}{10^{2}}$ for indoor (outdoor) localization. Dropout is applied after the second layer in both branches, with probability 0.2. 
We train the network once for position and once for orientation, using the  the loss function defined in Eq. \ref{eq1_pose_loss}. For indoor localization, we train once with $s_{x} = 1, s_{q} = 0$ and once with  $s_{x} = 0, s_{q} = 1$, for learning to regress the position and orientation, respectively. For outdoor localization, we observe that positional constraints are required when learning to regress the orientation and thus train with  $s_{x} = 1, s_{q} = 250$. For regressing the position, we follow the same strategy as for indoor localization. As oppose to other APRs using this type of loss, we do not fine tune  $s_{x}$ and $s_{q}$ per scene. We initialize the network with He initialization \cite{he2015delving} and train it for $300$ epochs on indoor scenes and for $1000$ epochs on outdoor scenes. We implement our model using PyTorch \cite{paszke2019pytorch} and train on a single NVIDIA Tesla V100 GPU.

\subsection{Results}
Tables \ref{tb:cambridge_res} and \ref{tb:7scenes_res} show the median position error (in meters) and orientation error (in degrees) of IRPNet and seven other APRs on the Cambridge Landmarks and the 7 Scenes dataset, respectively. Tables \ref{tb:cambridge_rank} and \ref{tb:7scenes_rank} give the average ranking for position and orientation, separately and combined, for each scene. The final ranking is taken as the average between the average position rank and average orientation rank per dataset. In this way, we avoid the bias associated with the different scales of position and orientation errors. 

For outdoor localization (Tables \ref{tb:cambridge_res} and \ref{tb:cambridge_rank}), we rank at the third place in terms of accuracy. The first and second most accurate APRs are variants of PoseNet, using the same architecture with different loss formulations. We note that IRPNet is the only model that consistently localizes in under $2m$ and $5^{\circ}$. Table \ref{tb:2m_5deg_comp} shows the percentage of pose error measurements under this threshold, which is closely related to medium accuracy bound.
We achieve the same ranking (third) also for indoor localization (Tables \ref{tb:7scenes_res} and \ref{tb:7scenes_rank}). In addition, IRPNet surpasses all other APRs in $28.6\%$ of the error measurements.

As shown in Tables \ref{tb:cambridge_rank} and \ref{tb:7scenes_rank}, together with PoseNet+Geometrical Loss, IRPNet is the only APR to consistently rank at the top-three on both benchmarks. Interestingly, the APRs ranked first for indoor (MapNet) and outdoor (PoseNet+Learnable), are not as accurate on both tasks. MapNet is ranked sixth for outdoor localization, while PoseNet+Learnable Loss is ranked at the fourth place.

In addition to accuracy, we also compare the training time of our network to a baseline APR (PoseNet with a ResNet-152 encoder). For a mini-batch of size 8, it took $140ms$ to train a batch with the baseline APR, and $12ms$ to train with IRPNet (over $10$ times faster). In addition, the input to our model is significantly smaller and constant in size. For example, when training on the 7 Scenes dataset, a typical APR  is trained with batches of $150K$ sized tensors (after image cropping), while IRPNet only requires a $4K$ sized input, without any initial image cropping.

\begin{table}[h!]
	\centering
	\caption{Percentage of pose error measurements under 2 meters and 5 degrees, for position and orientation respectively, on the Cambridge Landmarks dataset.}
	\label{tb:2m_5deg_comp}
	\begin{tabular}{|c||c|c|}
		\hline
		Method & Cambridge \\	
		\hline
		PoseNet (Learnable) & $87.5\%$ \\
		\hline
		GeoPoseNet & $87.5\%$ \\
		\hline
		\textbf{IPRNet (Ours)} & $100\%$ \\
		\hline
		MapNet & $87.5\%$ \\
		\hline
		SVS-Pose & $62.5\%$ \\
		\hline
		LSTM-PN & $75\%$ \\
		\hline
		GPoseNet & $50\%$ \\
		\hline
		BayesianPN & $37.5\%$ \\
		\hline
		PoseNet & $25\%$ \\
		\hline
		IR Baseline & $75\%$ \\
		\hline
	\end{tabular}
\end{table}

Using encodings trained for visual similarity also allows an easy integration into existing localization pipelines, such as \cite{taira2018inloc}. The trained model can provide a first rough estimate, serving as another supporting route. As oppose to encodings learned with a pose encoder, IR encodings carry visual similarity information. This can explain, why we were able to independently train our model for orientation and position, while other APRs were not able to successfully adopt this strategy \cite{kendall2015posenet,kendall2017geometric}.

\section{Conclusion}
Current APRs rely on a scene-specific encoder to generate a global pose feature vector, which is mapped to an estimated pose vector using an MLP regressor. In this work, we explore an alternative paradigm, where instead of training both an encoder and a pose regressor per scene, we train only the regressor and use a pre-trained visual IR encoder. This paradigm allows for shorter training time, lighter storage which does not depend on a scene-specific model and for integration into existing localization pipelines which are already using IR encodings. We propose an MLP architecture, named IRPNet, for evaluating this paradigm and show that it is able to achieve consistent localization (third place on both outdoor and indoor benchmarks) with minimal tuning of the loss function. Despite its simplicity, this architecture is also the only APR to consistently achieve medium accuracy on an outdoor benchmark.
The fact that (scene/dataset) agnostic visual encodings were sufficient to competitively regress the camera pose, compared to scene-specific pose encodings, suggests that APRs could benefit from other independent representations, coming from other modalities. For example, using pre-trained models for image segmentation or for extracting local visual representations.  More research into regressor architectures is also required to allow for integrating multi-modality information about the underlying map, capturing both visual and spatial constraints. Finally, given the multi-objective nature of the problem, separately training for orientation and position could take advantage of different modalities that are specific or more informative for a given objective. 

\bibliographystyle{IEEEtran}
%

\end{document}